\documentclass[conference]{IEEEtran}
\IEEEoverridecommandlockouts
\usepackage{cite}
\usepackage{fourier} 
\usepackage{amsmath,amssymb,amsfonts}
\usepackage{algorithmic}
\usepackage{graphicx}
\usepackage{textcomp}
\usepackage{xcolor}
\usepackage{adjustbox}

\usepackage{booktabs}

\def\BibTeX{{\rm B\kern-.05em{\sc i\kern-.025em b}\kern-.08em
    T\kern-.1667em\lower.7ex\hbox{E}\kern-.125emX}}
\begin{document}
\bstctlcite{IEEEexample:BSTcontrol}

\title{RAMOTS: A Real-Time System for Aerial Multi-Object Tracking based on Deep Learning and Big Data Technology}
\author{
\IEEEauthorblockN{Nhat-Tan Do\IEEEauthorrefmark{1}\IEEEauthorrefmark{2}, Nhi Ngoc-Yen Nguyen\IEEEauthorrefmark{1}\IEEEauthorrefmark{2}, Dieu-Phuong Nguyen\IEEEauthorrefmark{1}\IEEEauthorrefmark{2}, and Trong-Hop Do\IEEEauthorrefmark{1}\IEEEauthorrefmark{2}}

\IEEEauthorblockA{\IEEEauthorrefmark{1}Faculty of Information Science and Engineering, University of Information Technology\\
\IEEEauthorrefmark{2}Vietnam National University, Ho Chi Minh City, Vietnam \\
Corresponding author(s). Email(s): \texttt{hopdt@uit.edu.vn} \\
Contributing authors: \texttt{\{21522575, 21521231, 21520091\}@gm.uit.edu.vn}
}
\thanks{These authors contributed equally to this work.} 
}

\maketitle

\begin{abstract}

Multi-object tracking (MOT) in UAV-based video is challenging due to variations in viewpoint, low resolution, and the presence of small objects. While other research on MOT dedicated to aerial videos primarily focuses on the academic aspect by developing sophisticated algorithms, there is a lack of attention to the practical aspect of these systems. In this paper, we propose a novel real-time MOT framework that integrates Apache Kafka and Apache Spark for efficient and fault-tolerant video stream processing, along with state-of-the-art deep learning models YOLOv8/YOLOv10 and BYTETRACK/BoTSORT for accurate object detection and tracking. Our work highlights the importance of not only the advanced algorithms but also the integration of these methods with scalable and distributed systems. By leveraging these technologies, our system achieves a HOTA of 48.14 and a MOTA of 43.51 on the Visdrone2019-MOT test set while maintaining a real-time processing speed of 28 FPS on a single GPU. Our work demonstrates the potential of big data technologies and deep learning for addressing the challenges of MOT in UAV applications.
\end{abstract}

\begin{IEEEkeywords}
Multi-Object Tracking, Unmanned aerial vehicle, Big data processing, Streaming processing, Real-time
\end{IEEEkeywords}

\section{Introduction}

The number of UAVs (Unmanned aerial vehicles) has explosively grown in the last decade, with diverse applications in fields such as agriculture, communications, delivering, navigation and surveillance \cite{yuanUltrareliableIoTCommunications2018, radoglou-grammatikisCompilationUAVApplications2020, yooDroneDeliveryFactors2018, macrinaDroneaidedRoutingLiterature2020}. With their capability to fly and stay airborne without human-in-the-loop operation, they are more cost-efficient and more reliable by removing potential human errors.

Multi-Object Tracking (MOT) is the task of detecting multiple objects in video and associating these detections over time with correct identities. As a foundational task in computer vision, The MOT challenge receives extensive research interests, with a significant number of researches and works dedicated to improving its performance. While MOT is generally a large and attractive problem in the research community, MOT research dedicated to UAV videos is relatively limited. Unlike fixed surveillance cameras, MOT in UAV videos presents unique challenges due to the dynamic nature of the scenes: constantly changing in altitude and perspective, small object sizes, and high-speed movements. The limited computational resources on small UAV platforms pose an additional challenge when applying complex deep learning models for real-time object tracking, making most promising MOT strategies proposed in general not applicable to UAV-specific MOT.

While other research on Multi-Object Tracking (MOT) dedicated to UAV videos primarily focuses on the academic aspect by developing sophisticated algorithms, there is a lack of attention to the practical aspect of these systems. In practice, UAVs typically collect data and stream it to servers, where most processing occurs. In our work, we address not only the algorithmic aspects but also emphasize the practicality of the system. The state-of-the-art deep learning algorithms alone can't ensure the system will run smoothly and practically, we ensure that the system can handle large volumes of data and deliver results in real-time efficiently by integrating the whole system with big data platforms, namely Apache Spark and Apache Kafka. Our work highlights the importance of advanced algorithms and integrating these methods with scalable and distributed systems. By leveraging these technologies, our system achieves a HOTA of 48.14 and a MOTA of 43.51 on the Visdrone2019-MOT test set while maintaining a real-time processing speed of 28 FPS on a single GPU. Our work demonstrates the potential of big data technologies and deep learning for addressing the challenges of MOT in UAV applications. 

In this work, we propose a novel real-time Multi-object Tracking approach on UAV Footage by integrating Apache Kafka, Apache Spark, and state-of-the-art Deep Learning models such as YOLOv8/YOLOv10 and BYTETRACK/BoTSORT. The critical contributions of our work can be summarized as follows:

\begin{itemize}
    \item  \textbf{Novel Architecture Integration}. We introduce an architecture that seamlessly integrates Apache Kafka, a distributed streaming platform, and Apache Spark, a powerful cluster computing framework. This innovative combination enables efficient ingestion, distribution, and parallel processing of high-volume video streams from multiple UAVs, ensuring reliable and fault-tolerant data delivery.

    \item \textbf{State-of-the-Art Multi-Object Tracking Algorithm}. At the core of our framework is a robust multi-object tracking algorithm that integrates advanced computer vision techniques with efficient data processing pipelines. This algorithm utilizes deep learning models trained on diverse datasets, enabling precise detection and tracking of various object types—including vehicles, pedestrians, and aerial targets—under challenging conditions such as occlusions, rapid movements, and appearance variations.

    \item \textbf{Scalable and Distributed Processing}. Our framework capitalizes on the distributed nature of Apache Kafka and Apache Spark, facilitating horizontal scaling to accommodate an increasing number of UAVs and video streams. Kafka's partitioning and replication mechanisms, combined with Spark's cluster computing capabilities, enable parallel processing of video frames across multiple compute nodes, thereby accelerating object detection and tracking tasks.

    \item \textbf{Real-Time Performance}. By harnessing big data technologies and optimized algorithms, our framework achieves real-time performance, delivering timely and accurate situational awareness for critical applications such as security, traffic monitoring, search and rescue operations, and environmental monitoring.

\end{itemize}

\section{Related Works} \label{related-works}

\textbf{Object Detection.} Object detection, a fundamental task in computer vision, plays a crucial role in any MOT system. Object detection has been made a significant process during the past decade with the advent of many deep learning approaches. 

Two-stage detectors, such as Fast R-CNN\cite{girshickFastRcnn2015} and Faster R-CNN\cite{shouxinrenFasterRCNNRealtime2015}, utilize region proposal networks to identify potential objects, followed by a classification and bounding box regression step. This method offers high accuracy but can be computationally expensive. One-stage detectors, like YOLO \cite{redmonYouOnlyLook2016, wangYolov10RealtimeEndtoend2024, jocherYOLOUltralytics2023} or SSD \cite{liuSsdSingleShot2016}, directly predict object classes and bounding boxes in a single pass. They prioritize speed and are suitable for real-time applications but may sacrifice some accuracy compared to two-stage detectors. Transformer-based architectures, like DETR, have also emerged for object detection, offering improved accuracy and the ability to handle complex scenes effectively. These advancements in object detection contribute significantly to enhancing the accuracy and robustness of MOT systems by providing reliable object detection as input.

\textbf{Data Association.} Many algorithms and strategies were developed to help MOT system match between tracklets and detection boxes. SORT \cite{bewleySimpleOnlineRealtime2016}, a Simple online and real-time tracking algorithm that utilizes location and motion cues in a very simple yet effective way. Many significant methods adopt SORT, with some modifications to achieve favorable performance \cite{zhangByteTrackMultiObjectTracking2022, zhangFairMOT2021, mingyangHybridSORTWeakCues2023, caoObservationcentricSortRethinking2023, wangJDE2020, yiUCMCTrackMultiObjectTracking2024, wojkeSimpleOnlineRealtime2017}. Using the location and motion cues only is not robust to scenes with low certainty, large camera motion, or low frame rate. In these cases, appearance cues appear to work more efficiently. In scenarios when an object is occluded for a brief moment or even for a long period of time, it can be recognized using appearance information. DeepSORT \cite{wojkeSimpleOnlineRealtime2017} replaced the simple association metric in SORT with a more informed metric that combines motion and appearance cues. MOTDT \cite{chenRealtimeMultiplePeople2018} matches tracklets with detection by using appearance information first, then using IoU to match the remaining tracklets. FairMOT \cite{zhangFairMOT2021} addresses the issue of imbalanced optimization in multi-object tracking. It uses a single network to estimate object detections and perform re-identification (ReID) simultaneously. By jointly learning these tasks, FairMOT improves feature representation for both detection and ReID, leading to more robust and accurate tracking, especially in crowded scenes. ByteTrack \cite{zhangByteTrackMultiObjectTracking2022} introduces a new data association method, BYTE, which utilizes every detection box, including those with low confidence scores. This strategy involves associating high-scoring detections with existing tracklets and then matching remaining low-scoring detections with unmatched tracklets. This approach is reportedly efficient in working with occlusion, motion blur, or size changes. BoTSORT \cite{aharonBoTSORTRobustAssociations2022} takes a step further by adding improvements on top of ByteTrack, namely camera motion compensation-based (CMC) features tracker, with a new simple yet effective method for IoU and ReID’s cosine-distance fusion. Additionally, although inspired by ByteTrack, BoTSORT incorporates a more sophisticated data association strategy to handle challenging cases like occlusions and missed detections. 

\textbf{MOT on UAV videos.} 
Multiple Object Tracking (MOT) in UAV videos presents unique challenges due to the small size, fast motion, and varying perspectives of targets. Early approaches, often adapted from ground-based MOT methods, struggled with the non-linear motion models and abrupt appearance changes common in UAV footage. The advent of deep learning has led to significant advancements, with two-stage detectors like YOLO \cite{redmonYouOnlyLook2016} and Faster R-CNN \cite{shouxinrenFasterRCNNRealtime2015} combined with ReID techniques \cite{3} showing promise. More recently, one-shot MOT methods that integrate detection and ReID into a single network have gained popularity due to their real-time capabilities \cite{liuUAVMOT2022}. However, challenges such as long-term occlusions, varying object densities, and adverse weather conditions remain. Future research is likely to explore online learning, motion prediction, and the integration of contextual information to further improve tracking performance in UAV videos.

\section{Proposed real-time multi-object tracking architecture}

\subsection{Proposed real-time UAV video multi-object tracking System}

\begin{figure*}[ht]
    \centering
    \includegraphics[width=\textwidth]{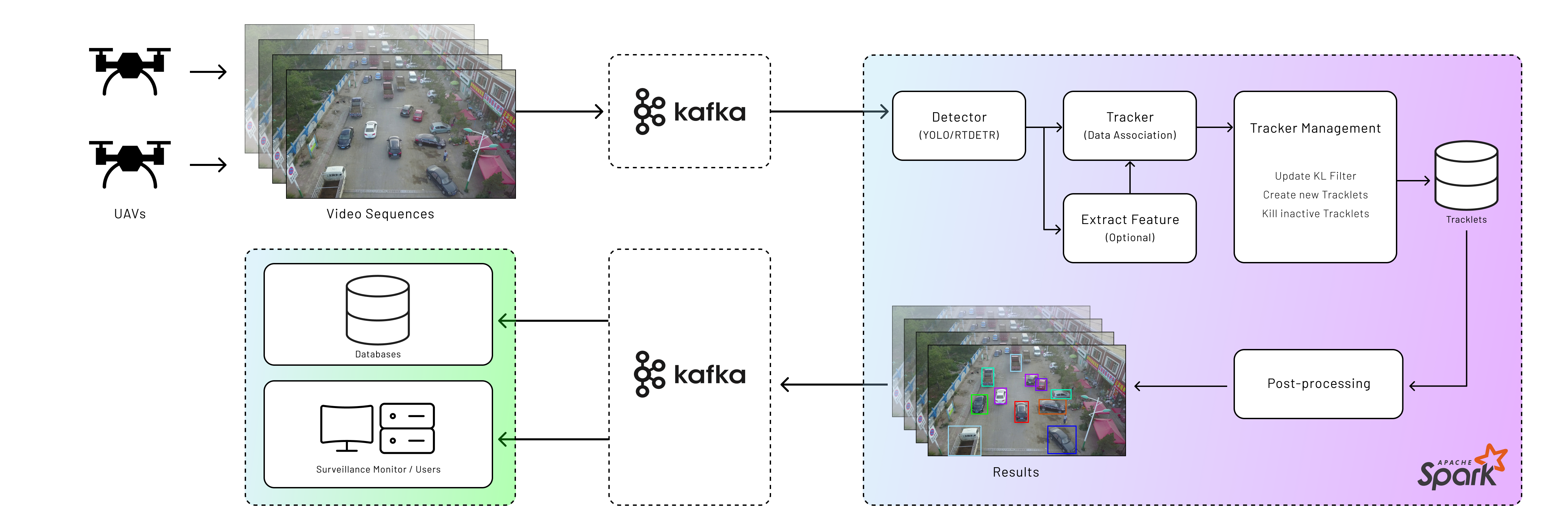}
    \caption{Our proposed Real-time UAV Videos Multi-object Tracking System}
    \label{fig:system}
\end{figure*}

Our system starts by getting video footage from UAVs then sent to the server via a Kafka Producer. The server's Consumer receives and processes the footage through the MOT framework in Apache Spark. The results are then sent back to the client via Kafka. The system's overall architecture is illustrated in Fig. \ref{fig:system}. 

\subsection{Integration of Apache Kafka and Apache Spark}

Kafka's architecture is tailored for \textit{high-throughput, fault-tolerant, and real-time data streaming}. This is critical for promptly processing and analyzing streaming video data to facilitate MOT in real time. 

Apache Kafka is used for real-time data ingestion. It serves as a message broker between the cameras and our processing server. The footage from multiple cameras is recorded and transmitted to the server via a Kafka Producer. The server's Consumer receives the footage and sends each frame to the Person Detection model sequentially. On the other hand, Apache Spark is used for real-time data processing: It receives the frames from Kafka, runs them through the MOT framework, and sends the results back to Kafka. The results are then sent back to the client.

In our system, we bring together Apache Kafka and Apache Spark to handle real-time data streaming and processing. This integration of Apache Kafka and Apache Spark allows our system to process and analyze video footage in real-time and can be easily scaled to handle large amounts of data.


\subsection{Multi-Object Tracking Algorithm}

The multi-object tracking algorithm can be divided into two main stages: Object Detection and Tracking. This section will explore how various state-of-the-art models are employed in each stage to achieve reliable and accurate multi-object tracking. By integrating these advanced object detection models and tracking algorithms, the multi-object tracking algorithm ensures high accuracy and robustness in diverse and challenging environments, maintaining consistent object identities across video frames. 

\subsubsection{Object Detection}

In the object detection stage, models such as YOLOv8, YOLOv10, RT-DETR, and Faster R-CNN are utilized. YOLOv8, a refined and optimized version of the YOLO family, is known for its balance between speed and accuracy, making it suitable for real-time applications where maintaining high frame rates is crucial. YOLOv10 builds on its predecessors with enhanced feature pyramids and attention mechanisms, providing precise object localization and classification in challenging environments. RT-DETR, or Real-Time Detection Transformer, leverages transformer architectures to capture long-range dependencies and the efficiency of convolutional networks for feature extraction, excelling in detecting small or overlapping objects in complex scenes. Additionally, Faster R-CNN, a widely used model in object detection, combines region proposal networks (RPN) with Fast R-CNN, allowing it to detect objects with high accuracy and robustness. Its two-stage approach, where the first stage generates region proposals and the second stage refines these proposals and performs classification, ensures precise detection even in complicated scenarios.

\subsubsection{Multi-Object Tracking}

The tracking stage involves associating detected objects across consecutive frames to maintain their identities over time, utilizing advanced algorithms such as ByteTrack, BoT-SORT, and SMILETrack-R. ByteTrack enhances robustness by integrating both high-confidence and low-confidence detections into a unified framework, employing a byte-level matching approach to effectively manage partial occlusions and varying object appearances. BoT-SORT builds on the classic SORT algorithm with improvements like adaptive Kalman filtering, IoU-based matching, and cascade matching strategies, ensuring more reliable object associations. SMILETrack-R introduces a novel method that incorporates re-identification (ReID) features alongside self-motivated interpolation and learning-based techniques, adeptly handling challenging scenarios such as abrupt motion changes and occlusions through dynamic trajectory updates. These methods not only reportedly result in SOTA performance in terms of both accuracy and performance, but they are also simple by nature and not specifically built for any particular dataset. This reason makes BYTETRACK and BoTSORT suitable for our system, which utilizes aerial footage from UAVs.

\section{Experiments} \label{experiments}


\begin{figure*}[ht]
    \centering
    \includegraphics[width=0.8\textwidth]{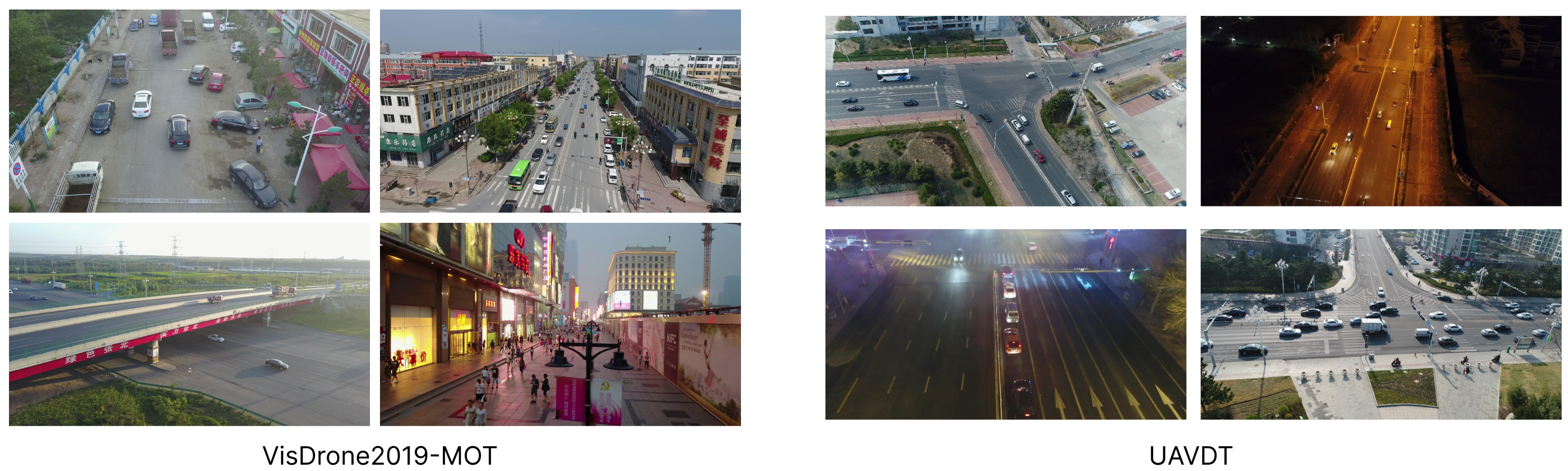}
    \caption{samples of UAVDT and VisDrone dataset. The videos from the two datasets cover various aspects, including location, environment, lighting, objects, and density.}
    \label{fig:dataset_samples}
\end{figure*}

\textbf{Datasets.} We conducted a fair evaluation on 2 largest publicly available datasets on MOT on UAV videos, namely VisDrone2019-MOT \cite{zhuVisDrone2021} and UAVDT \cite{duUAVDT2018}. VisDrone2019 is a large-scale benchmark for various important computer vision tasks, including Object Detection (OD), Single Object Tracking (SOT), Multi-Object Tracking (MOT), and Crowd Counting. The benchmark dataset consists of 288 video clips formed by 261,908 frames and 10,209 static images captured by various drone-mounted cameras, covering a wide range of aspects including location (taken from 14 different cities separated by thousands of kilometers in China), environment (urban and country), objects (pedestrian, vehicles, bicycles, etc.), and density (sparse and crowded scenes) \cite{zhuVisDrone2021}. UAVDT is also a multi-task large-scale UAV benchmark (i.e., about 80,000 representative frames from 10 hours of raw videos) for OD, SOT, and MOT. UAVDT consists of 100 video sequences, which are selected from over 10 hours of videos taken with a UAV platform at a number of locations in urban areas, representing various common scenes, including squares, arterial streets, toll stations, highways, crossings, and T-junctions. The videos are recorded at 30 frames per second (fps), with a JPEG image resolution of 1080 × 540 pixels \cite{duUAVDT2018}. VisDrone2019 benchmark covers a broader range of object classes than UAVDT. While UAVDT focuses solely on vehicles, VisDrone2019 includes a wider and more detailed set of classes, such as cars, motorcycles, and pedestrians. Sample of UAVDT and VisDrone dataset are shown in Figure \ref{fig:dataset_samples}.

\textbf{Metrics.} For the evaluation of performance, we use MOTA (Multiple Object Tracking Accuracy), IDF1 (Identification F1 Score), and IDs (Identity Switches) from the CLEAR metrics \cite{bernardinEvaluatingMultipleObject2008} along with a newly proposed metric, HOTA \cite{luitenHotaHigherOrder2021}. MOTA is computed based on false positives, false negatives, and identity switches, indicating how well the tracker maintains accurate object identities over time. IDF1 focuses on the accuracy of object association, which reflects the balance between correctly linking detections to tracked objects and creating new tracks for untracked objects. IDs indicates the number of times the tracker incorrectly "switches" the identity of an object being tracked to another identity. CLEAR metrics, although widely used in the research community, are still limited by their imbalance nature: while MOTA focuses more on the detection performance, IDF1 focuses more on the association performance. HOTA is a comprehensive metric that aims to capture the overall accuracy of an MOT system by considering three key aspects, including (1) detection accuracy, (2) association accuracy, and (3) localization accuracy. Fast inference is also a crucial part of real-time MOT system. We evaluate the speed of our methods by calculating FPS (Frame per second), which indicates how fast our system can run in one second. 

\textbf{Implementation Details.} For a fair comparison, all detection methods were trained on the VisDrone2019-DET train set using YOLOv8, YOLOv10, and RT-DETR architectures. The models were pre-trained on the COCO dataset and fine-tuned for 4 epochs with a batch size of 16 and an initial learning rate of 0.01. Data augmentation techniques, such as random horizontal flips and scaling, were applied during training. We utilized a single NVIDIA Tesla A100 GPU for training and inference.

In the inference phase, for both BoTSORT and BYTETRACK, we use the detection score threshold $\tau = 0.6$. If the IoU between the detection box and the tracklet box is smaller than 0.2 in the linear assignment step, we will reject the matching. Lost tracklets will be kept for 30 frames (\textit{track buffer}), in case they reappear. All inferences are run on the same machine with 16GB of RAM, $\sim$3.2GHz CPU, and an NVIDIA RTX 3060 GPU.


\subsection{Results and Discussion}

\begin{table*}[ht]
\setlength{\tabcolsep}{4pt} 
\caption{Comparative analysis of Classical MOT methods (without Detector-Tracker Combination). Results Aggregated from Literature on VisDrone2019-MOT and UAVDT dataset}\label{tab:exp-results}
\begin{adjustbox}{width=\textwidth}
\begin{tabular*}{\textheight}{@{\extracolsep\fill}lrrrrrrrrr}
\toprule%
                                              & \multicolumn{4}{@{}c@{}}{VisDrone2019-MOT}                         & \multicolumn{4}{@{}c@{}}{UAVDT}& \multicolumn{1}{@{}c@{}}{} \\\cmidrule{2-5}\cmidrule{6-9}%
Methods                                       & HOTA$\uparrow$ & MOTA$\uparrow$ & IDF1$\uparrow$ & IDs$\downarrow$ & HOTA$\uparrow$& MOTA$\uparrow$& IDF1$\uparrow$& IDs$\downarrow$ & FPS$\uparrow$ \\
\midrule
MOTDT \cite{chenRealtimeMultiplePeople2018}  & -              & -0.80          & 21.60          & 1437            & -             & -             & -             & -              & -          \\
SORT \cite{bewleySimpleOnlineRealtime2016}    & -              & 14.00          & 38.00          & 3629            & -             & 39.00         & 43.70         & 2350           & -          \\
JDE \cite{wangJDE2020}                        & -              & 26.60          & 34.90          & 3200            & -             & 39.50         & 55.30         & 3124           & \textbf{17.8}       \\
FairMOT \cite{zhangFairMOT2021}               & -              & 30.80          & 41.90          & 3007            & -             & 44.90         & 60.90         & 2279           & 18         \\
MOTR \cite{zengMotrEndtoendMultipleobject2022}& -              & 22.80          & 41.40          & \textbf{959}   & -             & -             & -             & -              & -          \\
TrackFormer \cite{meinhardtTrackformer2022}   & -              & 25.00          & 30.50          & 4840            & -             & -             & -             & -              & -          \\
FPUAV \cite{wuFPUAV2022}                      & -              & 34.30          & 45.00          & 2138            & -             & -             & -             & -              & 17.6       \\
UAVMOT  \cite{liuUAVMOT2022}                  & -              & \textbf{36.10}          & \textbf{51.00}          & 2775            & -             & \textbf{48.6}          & \textbf{66.2}          & \textbf{1999}           & -          \\
\bottomrule
\end{tabular*}
\end{adjustbox}
\end{table*}

\begin{table*}[ht]
\setlength{\tabcolsep}{4pt} 
\caption{Comparison of different methods with combined Detector and Tracker under the VisDrone2019-MOT and UAVDT dataset. The best results are shown in \textbf{bold}}\label{tab:exp-results_2}
\begin{adjustbox}{width=\textwidth}
\begin{tabular*}{\textheight}{@{\extracolsep\fill}lrrrrrrrrr}
\toprule%
                                              & \multicolumn{4}{@{}c@{}}{VisDrone2019-MOT}                         & \multicolumn{4}{@{}c@{}}{UAVDT}& \multicolumn{1}{@{}c@{}}{} \\\cmidrule{2-5}\cmidrule{6-9}%
Methods                                       & HOTA$\uparrow$ & MOTA$\uparrow$ & IDF1$\uparrow$ & IDs$\downarrow$ & HOTA$\uparrow$& MOTA$\uparrow$& IDF1$\uparrow$& IDs$\downarrow$ & FPS$\uparrow$ \\
\midrule

YOLOv8l, ByteTrack                           & 43.59          & 41.77          & 54.78          & 4212            & 53.46         & 34.62         & 70.32         & 786            & 28         \\
YOLOv8l, BoTSORT                             & \textbf{48.14} & \textbf{43.51} & \textbf{61.46} & 3118            & 53.93         & 33.86         & 71.11         & 725            & 28         \\
YOLOv8l, SMILETrack-R                        & 40.67          & 12.90          & 48.39          & 22412           & -             & -             & -             & -              & 5           \\
YOLOv8x, BoTSORT                             & 43.90          & 42.29          & 55.14          & 4432            & 53.42         & 30.69         & 70.21         & 756            & 28         \\
YOLOv10l, ByteTrack                          & 35.43          & 26.66          & 38.85          & \textbf{1788}            & 56.48         & \textbf{53.87}& 74.60& \textbf{576}   & 24         \\
YOLOv10l, BoTSORT                            & 44.33          & 39.40          & 55.68          & 2044            & \textbf{57.21}& 53.64         & \textbf{75.53}         & 600            & 24         \\
RT-DETR-l, ByteTrack                         & 41.14          & 36.34          & 50.32          & 3600            & 50.25         & 8.75          & 63.54         & 788            & 21         \\

RT-DETR-l, BoTSORT                           & 41.77          & 35.65          & 50.32          & 3607            & 50.50         & 8.51          & 63.86         & 795            & 21         \\

\bottomrule
\end{tabular*}
\end{adjustbox}
\end{table*}

\begin{figure}
    \centering
    \includegraphics[width=8cm]{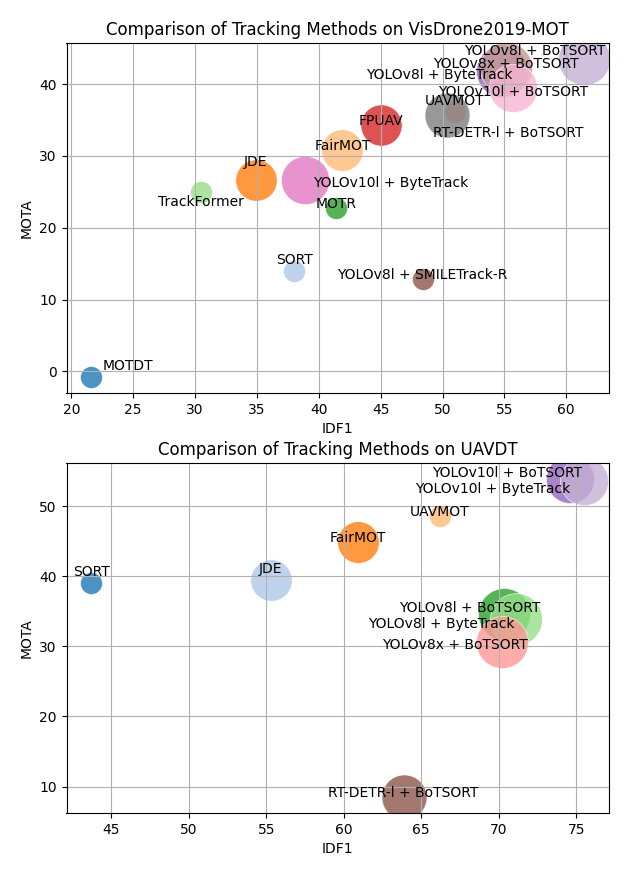}
    \caption{Comparative performance of Deep Learning-based MOT methods on VisDrone2019-MOT and UAVDT dataset. The horizontal axis, vertical axis, and radius of the circle are IDF1, MOTA, and FPS, respectively. Detail comparison are shown in \ref{tab:exp-results} and \ref{tab:exp-results_2}}
    \label{fig:results}
\end{figure}

The results of our experiments are shown in Table \ref{tab:exp-results}, \ref{tab:exp-results_2} and are visualized in Figure \ref{fig:results}. $\uparrow$/$\downarrow$ indicate that higher/lower is better, respectively. The best results are shown in \textbf{bold}. The YOLOv8l and YOLOv10l architectures, particularly when paired with the BoTSORT tracker, consistently demonstrated superior performance across various accuracy metrics, with $48.14\%$ HOTA on VisDrone2019-MOT with YOLOv8l+BoTSORT and $57.21\%$ on UAVDT with YOLOv10l+BoTSORT, outperformed other combinations by a large margin. This suggests their effectiveness in accurately detecting and maintaining consistent tracks of objects in aerial videos, even in challenging scenarios.

While both BoTSORT and ByteTrack proved to be capable trackers, BoTSORT often edged out ByteTrack in terms of accuracy about $2\%$ to $5\%$ on the same setting, highlighting its potential for more robust and reliable tracking over time.  The newer DETR-based methods, specifically RT-DETR-l, showed promise, especially on the UAVDT dataset. However, their performance on the more complex VisDrone2019-MOT dataset was less consistent, indicating a need for further optimization to enhance their generalization capabilities.


It is noticeable that YOLOv10l performed much worse than these combinations of YOLOv8l (YOLOv8l performed better by $8.16\%$ and $3.81\%$ HOTA on VisDrone2019 when combined with ByteTrack and BoTSORT respectively), its IDs score is the lowest ($1788$ compares to $4212$ of YOLOv8l with the same setting, which is $42.4\%$ of the later). This means that YOLOv10l has the least number of IDs, or in simple words, it managed to "follow" objects well without mistakenly assigning new IDs. This insight, along with its highest score in the UAVDT dataset, indicates that YOLOv10l's combinations are generalized well and can perform in different scenarios more effectively.

In conclusion, deep learning-based approaches, notably those leveraging YOLO architectures and the BoTSORT tracker, hold significant promise for advancing multi-object tracking in aerial videos. However, the quest for real-time, highly accurate, and generalizable MOT systems remains an active area of research, with ongoing efforts focused on optimizing algorithms, addressing dataset-specific challenges, and striking an optimal balance between speed and accuracy.


\section{Conclusion}

In conclusion, this study has explored the critical challenge of multi-object tracking (MOT) in UAV videos, a domain characterized by dynamic scenes, small object sizes, and limited computational resources. Other research on Multi-Object Tracking dedicated to UAV videos primarily focuses on the academic aspect by developing sophisticated algorithms but lacks attention to the practical aspect of these systems. In our work, we address not only the algorithmic aspects but also emphasize the practicality of the system. We introduced a novel framework that harnesses the power of big data technologies, including Apache Kafka and Apache Spark, to enable scalable and efficient processing of video streams from multiple UAVs. By integrating a state-of-the-art MOT algorithm based on deep learning models and optimized data pipelines, our framework achieves real-time performance, delivering accurate and timely situational awareness for various applications. Our system achieves a HOTA of 48.14 and a MOTA of 43.51 on the Visdrone2019-MOT test set while maintaining a real-time processing speed of 28 FPS on a single GPU. Our work demonstrates the potential of big data technologies and deep learning for addressing the challenges of MOT in UAV applications. 

Through extensive experiments on the VisDrone2019-MOT and UAVDT datasets, we demonstrated the superior performance of our framework compared to existing methods, particularly those leveraging YOLO architectures and the BoTSORT tracker. Our findings emphasize the effectiveness of deep learning-based approaches for MOT in aerial videos while also highlighting the challenges posed by complex datasets like VisDrone2019-MOT, which demand further optimization to improve generalization capabilities.

Looking ahead, we envision several promising directions for future research. These include exploring more advanced deep learning models, incorporating additional data sources such as LiDAR and radar, and investigating strategies for adaptive resource allocation in UAV swarms. Ultimately, our goal is to develop robust, real-time MOT systems that empower UAVs to effectively navigate and interact with their environment, ultimately benefiting a wide range of industries and applications.

\bibliographystyle{abbrv}
\bibliography{bibliography}

\end{document}